\pdfoutput=1

\documentclass[11pt]{article}

\usepackage[preprint]{acl}

\usepackage{times}
\usepackage{latexsym}

\usepackage[T1]{fontenc}

\usepackage[utf8]{inputenc}

\usepackage{microtype}

\usepackage{inconsolata}

\usepackage{siunitx}
\usepackage{caption}
\usepackage{subcaption}
\usepackage{graphicx}
\usepackage[most]{tcolorbox}
\usepackage{amsmath}
\usepackage{booktabs,arydshln}
\usepackage{multirow}
\usepackage{makecell}
\usepackage[compact]{titlesec}

\newenvironment{bracket}{}{}

\makeatletter
\def\adl@drawiv#1#2#3{%
        \hskip.5\tabcolsep
        \xleaders#3{#2.5\@tempdimb #1{1}#2.5\@tempdimb}%
                #2\z@ plus1fil minus1fil\relax
        \hskip.5\tabcolsep}
\newcommand{\cdashlinelr}[1]{%
  \noalign{\vskip\aboverulesep
           \global\let\@dashdrawstore\adl@draw
           \global\let\adl@draw\adl@drawiv}
  \cdashline{#1}
  \noalign{\global\let\adl@draw\@dashdrawstore
           \vskip\belowrulesep}}
\makeatother

\title{Rethinking Data Selection for Supervised Fine-Tuning}

\author{
Ming Shen   \\
Arizona State University \\
\texttt{mshen16}@asu.edu              \\
}

\begin{document}
\maketitle

\begin{abstract}

Although supervised finetuning (SFT) has emerged as an essential technique to align large language models with humans, it is considered superficial, with style learning being its nature. 
At the same time, recent works indicate the importance of data selection for SFT, showing that finetuning with high-quality and diverse subsets of the original dataset leads to superior downstream performance. 
In this work, we rethink the intuition behind data selection for SFT.
Considering SFT is superficial, we propose that essential demonstrations for SFT should focus on reflecting human-like interactions instead of data quality or diversity. 
However, it is not straightforward to directly assess to what extent a demonstration reflects human styles. 
Towards an initial attempt in this direction, we find selecting instances with long responses is surprisingly more effective for SFT than utilizing full datasets or instances selected based on quality and diversity.
We hypothesize that such a simple heuristic implicitly mimics a crucial aspect of human-style conversation: detailed responses are usually more helpful.

\end{abstract}
\section{Introduction}
\label{sec:intro}

Large language models (LLMs) \cite{NEURIPS2020_1457c0d6, Chowdhery2022PaLMSL, Anil2023PaLM2T, Touvron2023LLaMAOA, Touvron2023Llama2O, Jiang2023Mistral7} have shown remarkable capabilities in canonical natural language understanding and generation tasks in recent years. To further teach LLMs to align with humans by following human instructions and interacting with humans in user-friendly manners, supervised finetuning, or instruction tuning \cite{NEURIPS2022_b1efde53, wang-etal-2023-self-instruct, alpaca, Peng2023InstructionTW, vicuna2023, Xu2023WizardLMEL, chatgpt, Achiam2023GPT4TR}, has been applied as an indispensable alignment step where LLMs are finetuned on large sets of instruction-response demonstrations either annotated by human or synthesized by power proprietary models. 

Despite the success of alignment through SFT, LIMA \cite{zhou2023lima} proposes the superficial alignment hypothesis that the models’ knowledge and abilities are learned mostly during pretraining, and SFT is all about style learning \cite{gudibande2023false} of formatting the response in a human-like manner. At the same time, recent works have been designing data selection strategies to filter SFT datasets, showing finetuning LLMs with a subset of the original dataset leads to superior instruction-following capabilities compared with utilizing the whole. Two major principles for designing existing selection strategies \cite{Chen2023MaybeO0, Lu2023InsTagIT, alpagasus, Bukharin2023DataDM, Du2023MoDSMD, Wu2023SelfEvolvedDD, Liu2023WhatMG} are based on the quality and diversity aspects of the data. However, although these two factors are highly significant for developing AI systems in the pre-LLM era, we question their relevance under the SFT scenario in the current era of LLMs, given the superficial nature of SFT.


Adhering to such nature of SFT, we hypothesize that data selection for SFT should focus on picking demonstrations that reflect human style the most. For example, a demonstration of teaching LLM that Paris is the capital of France is unhelpful since LLM already gained such knowledge during the pretraining phase. 
On the contrary, demonstrations with human-like responses containing words such as "thanks" and "first of all", and structured responses containing numbered lists are helpful for SFT \cite{Lin2023TheUS}. 
However, directly identifying such demonstrations is not a straightforward process; thus, we investigate a simple heuristic by selecting demonstrations with long responses, inspired by the observation that if a response meets the instruction requirement in the first place, a longer version with more details is deemed more helpful. Two illustrative examples are given in Appendix Table \ref{tab:gpt4_explaination_1} and \ref{tab:gpt4_explaination_2} where both human and GPT-4 prefer the longer response with more details. Surprisingly, even such a simple heuristic leads to models with strong instruction-following capabilities.  
For Alpaca dataset \cite{alpaca}, finetuning LLaMA-2-7B \cite{Touvron2023Llama2O} with only top 1K instances with long responses leads to an average win rate of 68\% versus 20\% when utilizing entire dataset, 63\% versus 20\% when utilizing same amount of instances selected based on the quality aspect, and 70\% versus 15\% when utilizing instances with high diversity instructions. 

Overall, starting from the potentially superficial nature of SFT, we question the existing data selection strategies based on quality and diversity aspects and propose to select true important instances from a perspective of mimicking human style. To make an initial effort in this direction, we show that even a naive heuristic leads to strong performances compared with baseline selection strategies. This also sheds light on better curating SFT datasets instead of only selecting from an existing data pool.

\section{Method}

Given pretrained LLM $\mathcal{M}$ and SFT dataset $\mathcal{D}$ of demonstrations $(x_i, y_i)$, where $x_i$ is an open-ended instruction and $y_i$ is the response that completes the instruction, denote $\mathcal{M}_\Delta$ as $\mathcal{M}$ finetuned with $\Delta$, we aim to select a subset $\mathcal{D}^{\prime} \in \mathcal{D}$ such that, $\left| \mathcal{D}^{\prime} \right| \ll \left| \mathcal{D} \right|$ and $\mathcal{M}_{\mathcal{D}^{\prime}}$ shows superior instruction-following capability than $\mathcal{M}_\mathcal{D}$ evaluated on some open-ended instruction test sets. 

We adopt a naive yet effective heuristic of selecting SFT data with long responses, mimicking human-like demonstrations with rich details in the response. Please refer to the Introduction Section \ref{sec:intro} for more details of our motivation. Specifically, we rank each instance $(x_i, y_i)$ in $\mathcal{D}$ based on its response token length $|y_i|$ and then select instances with the top longest responses to form $\mathcal{D}^{\prime}$.

\section{Experiments}

\noindent \textbf{Training Setup.} \hspace{4pt} 
We investigate three SFT datasets for finetuning: Alpaca 52K \cite{alpaca}, WizardLM 70K \cite{Xu2023WizardLMEL}, and Dolly 15K \cite{DatabricksBlog2023DollyV2}. 
Please refer to Appendix Section \ref{sec:datasets_details} for details of the datasets.
We use LLaMA-2-7B \cite{Touvron2023Llama2O} as pre-trained LLM. We keep all training hyperparameters the same for fair comparisons. 
Please refer to Appendix Section \ref{sec:training_details} for training details.

\noindent \textbf{Baselines.} \hspace{4pt}
We compare against utilizing the entire dataset, random selection, and data selection strategies based on quality and diversity. 
For quality-based selection, we adopt the method proposed in \citet{alpagasus} to first directly prompt ChatGPT as an auto-grader to rate each instance in $\mathcal{D}$, then select instances with top scores as high-quality instances. Please refer to Appendix Section \ref{sec:quality_selection} for more details.
For diversity-based selection, we adopt K-Means-Clustering to select diversified instances on the embedding space of all training instructions. We use LLaMA-2-7B as the sentence encoder and take the average of the last hidden layer outputs over all tokens as sentence embedding. We cluster the entire dataset into 100 clusters and randomly sample an equal number of instances from each cluster. We also experiment with more advanced K-Centering-Greedy \cite{sener2018active} approach to select diverse instructions; however, it underperforms K-Means Clustering during downstream evaluation.

\noindent \textbf{Evaluation Setup.} \hspace{4pt}
To comprehensively evaluate instruction-following capabilities, we utilize test sets from 5 works: Self-Instruct \cite{wang-etal-2023-self-instruct}, Vicuna \cite{vicuna2023, zheng2023judging}, Koala \cite{koala_blogpost_2023}, WizardLM \cite{Xu2023WizardLMEL}, and LIMA \cite{zhou2023lima}. Appendix Table \ref{tab:testsets_stats} shows statistics details.

Given two systems' responses to an instruction, we perform a pairwise evaluation by prompting GPT-4 to assign a score from 1-10 with explanations for each response \cite{dubois2023alpacafarm, zheng2023judging, vicuna2023}. To mitigate the position bias \cite{Wang2023LargeLM, zheng2023judging}, we swap the position of two responses to let GPT-4 judge twice and define that a system wins or loses only when its response is preferred or dispreferred by GPT-4 during both pass; otherwise, it is a tie. Another crucial potential bias is verbosity bias, meaning LLM judges usually prefer longer responses. However, GPT-4 as judge mitigates this bias to large extent: \citet{zheng2023judging} shows that GPT-4 is robust to repetitive attack with failure rate of only 8\%; whereas other LLM judges such as ChatGPT and Claude \cite{claude} fail more than 90\% of the time.
Following \citet{alpagasus}, we use the evaluation prompt shown in Appendix Figure \ref{fig:gpt-prompt}) and winning score of $\frac{\text{\#Win}-\text{\#Lose}}{\text{\#Testset}} + 1$ where $\text{\#Testset} = \text{\#Win} + \text{\#Tie} + \text{\#Lose}$ as evaluation metric.

\begin{table}[!t]
    \centering
    \scalebox{0.8}{
        \begin{tabular}{@{}ccccc@{}}
        \toprule
        \textbf{Dataset} & \textbf{Size} & \textbf{Strategy} & \textbf{vs. Full} & \textbf{vs. Random} \\ \midrule
        \multirow{6}{*}{\textbf{\begin{tabular}[c]{@{}c@{}}Alpaca\end{tabular}}} & \multirow{3}{*}{1K} & Diversity & 1.0194 & 0.8913 \\
         &  & Quality & 1.1913 & 1.0961 \\
         &  & Longest & \textbf{1.4505} & \textbf{1.4631} \\ \cmidrule(l){2-5} 
         & \multirow{3}{*}{9K} & Diversity & 1.2058 & 1.0398 \\
         &  & Quality & 1.3058 & 1.1573 \\
         &  & Longest & \textbf{1.4718} & \textbf{1.3563} \\ \midrule
        \multirow{3}{*}{\textbf{\begin{tabular}[c]{@{}c@{}}WizardLM\end{tabular}}} & \multirow{3}{*}{6\%} & Diversity & 1.0748 & 1.0107 \\
         &  & Quality & 1.1641 & 1.0893 \\
         &  & Longest & \textbf{1.1942} & \textbf{1.1398} \\ \midrule
        \multirow{6}{*}{\textbf{\begin{tabular}[c]{@{}c@{}}Dolly\end{tabular}}} & \multirow{3}{*}{1K} & Diversity & 0.9078 & 1.1068 \\
         &  & Quality & 0.7796 & 0.9777 \\
         &  & Longest & \textbf{1.1379} & \textbf{1.2806} \\ \cmidrule(l){2-5} 
         & \multirow{3}{*}{3K} & Diversity & 0.8786 & 0.9845 \\
         &  & Quality & 1.0680 & 1.1583 \\
         &  & Longest & \textbf{1.1495} & \textbf{1.2466} \\ \bottomrule
        \end{tabular}
    }

    \caption{Winning scores of comparing data selection based on diversity, quality, and response length versus utilizing full dataset and random selection.}
    \label{tab:vs_full_random}
\end{table}

\begin{table}[!t]
    \centering
    \scalebox{0.8}{
        \begin{tabular}{@{}cccc@{}}
        \toprule
        \textbf{Dataset} & \textbf{Strategy-Size} & \textbf{vs. Diversity} & \textbf{vs. Quality} \\ \midrule
        \multirow{2}{*}{\textbf{\begin{tabular}[c]{@{}c@{}}Alpaca\end{tabular}}} & Longest-1K & 1.5311 & 1.4146 \\
         & Longest-9K & 1.3796 & 1.2883 \\ \midrule
        \textbf{\begin{tabular}[c]{@{}c@{}}WizardLM\end{tabular}} & Longest-6\% & 1.1476 & 1.0592 \\ \midrule
        \multirow{2}{*}{\textbf{\begin{tabular}[c]{@{}c@{}}Dolly\end{tabular}}} & Longest-1K & 1.1981 & 1.2854 \\
         & Longest-3K & 1.2942 & 1.1350 \\ \bottomrule
        \end{tabular}
    }

    \caption{Winning scores of selecting instances with long responses versus selection based on diversity and quality.}
    \label{tab:vs_baselines}
\end{table}

\section{Main Results}

For Alpaca dataset, we study selection with 1K and 9K instances; for Dolly, we study selection with 1K and 3K instances; for WizardLM, we select 6\% instances of the entire data. 

\noindent \textbf{Versus Utilizing Full Datasets.} \hspace{4pt} 
As shown in the penultimate column of Table \ref{tab:vs_full_random}, for Alpaca and Dolly, models finetuned on merely top 1k instances with long responses outperform utilizing full datasets with winning scores of 1.4505 and 1.1379. The same trend is observed when selecting top 9K and 3K instances, with winning scores of 1.4718 and 1.1495.
For WizardLM, we first also utilize top 1K instances; however, it underperforms the model finetuned on full dataset with winning score smaller than 1. We then directly increase selection size to 6\% (4,200), leading to a winner score of 1.1942. 
Overall, this demonstrates the effectiveness of selecting SFT instances based on response length, mimicking the human-like style of responses being detailed.

\noindent \textbf{Versus Baseline Strategies.} \hspace{4pt}
As shown in the last column of Table \ref{tab:vs_full_random}, selection with long responses leads to winning scores larger than 1 versus random selection in all comparisons. More importantly, as shown in Table \ref{tab:vs_baselines}, selection with long responses leads to winning scores larger than 1 versus quality-based and diversity-based selection in all comparisons. The highest score is 1.5311, compared to selecting top 1K high-diversity instances for Alpaca. The lowest score is 1.0592, compared to selecting top 6\% high-quality instances for WizardLM. 
Overall, the results support our hypothesis that selecting SFT data reflecting human-like styles is more important than recent popular selection strategies based on data quality and diversity.

\begin{table*}[!th]
    \centering
    \scalebox{0.76}{
        \begin{tabular}{@{}cccccccccccc@{}}
        \toprule
        & \textbf{LongForm} & \textbf{ELI5} & \textbf{ARC} & \textbf{HellaSwag} & \textbf{MMLU} & \textbf{TruthfulQA} & \textbf{Winogrande} & \textbf{GSM8K} & \textbf{BBH} & \textbf{\textit{AVG1}} & \textbf{\textit{AVG2}} \\ 
        \midrule
        Base & 14.62 & 11.51 & 50.17 & 72.02 & 45.65 & 38.80 & 73.80 & 11.50 & 39.63 & 13.07 & 47.37 \\
        \cdashlinelr{1-12}
        Full & 21.80 & 14.70 & 50.00 & 73.49 & 44.04 & 42.14 & 69.93 & 7.00 & 36.76 & 18.25 & 46.19 \\
        Random & 22.91 & 15.05 & 53.67 & 74.54 & 44.80 & 42.54 & 72.53 & 12.00 & 35.19 & 18.98 & 47.90 \\
        Diversity & 20.47 & 14.29 & 52.82 & 74.22 & \textbf{46.00} & 42.03 & 73.24 & \textbf{14.50} & 31.85 & 17.38 & 47.81 \\
        Quality & 22.39 & 15.85 & 54.18 & 74.36 & 44.80 & 41.43 & 73.24 & 13.00 & 35.74 & 19.12 & 48.11 \\
        Longest & \textbf{24.01} & \textbf{19.07} & \textbf{54.86} & \textbf{75.39} & 45.70 & \textbf{43.65} & \textbf{73.88} & \textbf{14.50} & \textbf{41.30} & \textbf{21.54} & \textbf{49.90} \\
        \bottomrule
        \end{tabular}
    }
    \caption{Performances on all benchmark datasets using base model, models finetuned with full Alpaca dataset, and subsets selected with various strategies. \textit{AVG1} is the average score for LongForm and ELI5, and \textit{AVG2} is the average score for the rest.}
    \label{tab:alpaca_benchmark}
\end{table*}

\noindent \textbf{More Discussions.} \hspace{4pt}
We find that quality-based and diversity-based data selections are less robust than selecting instances with long responses across different SFT datasets and selection sizes. For example, as shown in Table \ref{tab:vs_full_random}, selecting top 1K instances from Alpaca with high diversity underperforms random selection with winning score of 0.8913; however, expanding selection size to 9K leads to winning score larger than 1. For Dolly, both selecting 1K instances based on diversity and quality underperform utilizing whole dataset with winning scores less than 1, and quality-based selection underperforms random selection. After expanding selection size to 3K, quality-based selection achieves winning scores larger than 1 versus random selection and utilizing full dataset; however, diversity-based selection leads to winning scores less than 1 in both cases. 
Selection with long responses leads to winning scores larger than 1 in all comparisons.
Finally, we show results of using K-Centering-Greedy instead of K-Means Clustering in Appendix Table \ref{tab:coreser}.

\section{Analysis}

\begin{table}[ht]
\centering
    \scalebox{0.8}{
        \begin{tabular}{@{}cccc@{}}
        \toprule
         & \textbf{Alpaca-1K} & \textbf{WizardLM-6\%} & \textbf{Dolly-1K} \\ \midrule
        vs. Full & 0.1544 & 0.1883 & 0.2126 \\ \bottomrule
        \end{tabular}
    }
    \caption{Winning scores of utilizing instances with short responses versus utilizing full dataset.} 
    \label{tab:long_vs_short}
\end{table}

\noindent \textbf{Effect of Short Responses.} \hspace{4pt}
To investigate the role of instances with short responses, we finetune with top 1K, 6\%, and 1K instances with shortest responses from the three datasets, respectively, to compare with utilizing the entire dataset. Table \ref{tab:long_vs_short} shows that finetuning on instances with short responses leads to best winning score of only 0.21, meaning model loses 79\% of all comparisons.


\noindent \textbf{Are Detailed Responses Generated?} \hspace{4pt}
We aim to investigate whether model finetuned on instances with long responses generates detailed responses through a small-scale human evaluation. We random sample 50 comparisons where utilizing top 1K instances with long responses from Alpaca outperforms utilizing the entire Alpaca dataset evaluated on the Self-Instruct test set and carefully inspect the explanations given by GPT-4 for its preferences. The total number of win-tie-lose under this comparison setting is 151-36-65. We find that out of the 50 response pairs, both responses in 33 pairs meet the requirement of the instruction to roughly the same extent; however, GPT-4 prefers the winner because it is "\textit{more detailed}", explained by GPT-4, and more importantly, we human annotator do find detailed responses in these 33 comparisons make the winner more helpful. We provide two examples in Appendix Tables \ref{tab:gpt4_explaination_1} and \ref{tab:gpt4_explaination_2}.

\noindent \textbf{Potential Evaluation Bias.} \hspace{4pt}
Although GPT-4 has shown to be robust to verbosity bias compared to other judges, such as ChatGPT and Claude \cite{zheng2023judging}, our study is vulnerable under this bias since our proposed method tends to generate long output. Thus, we first perform a small-scale human analysis again from the sampled 50 response pairs in the last section and identify only 4 cases when a response is of lower quality but still preferred by GPT-4 because it is "\textit{more detailed}", explained by GPT-4. This 8\% false positive rate showcases that GPT-4's evaluation is reliable enough.

Then, we proceed to explicitly instruct GPT-4 not to favor long responses, hoping to eliminate verbosity bias further. We modify the evaluation prompt in Figure \ref{fig:gpt-prompt}: we remove the rating criteria of "\textit{level of details}" and explicitly add "\textit{Do not allow the length of the responses to influence your evaluation}". We use the new prompt to perform comparisons between utilizing top 1K instances with long responses versus entire Alpaca data on all test sets, and we find an agreement rate of 94.17\% against using the original prompt. This means our evaluation is not biased towards prompt choice.

\noindent \textbf{Long-Response Instructions Distribution.} \hspace{4pt}
To study the distribution of instructions with long responses, we use Berkeley Neural Parser \cite{kitaev-klein-2018-constituency, kitaev-etal-2019-multilingual} to parse the top 20 most common root verbs and their top 4 direct noun objects of the instructions of top 1K instances with long responses from Alpaca dataset following \citet{wang-etal-2023-self-instruct}. We find that instruction root verbs "\textit{write}", "\textit{generate}", "\textit{create}", and "\textit{compose}" compose 70\% of all instructions. 
Please refer to Appendix Section \ref{sec:distribution} for a direct view of verb-noun structures for all datasets.

\noindent \textbf{Benchmark Performances.} \hspace{4pt}
Besides open-ended instruction-following capability, we also evaluate various benchmark datasets. One benefit of finetuning on instances with long responses is that the model can generalize well on long-form text generation tasks. Thus we evaluate on LongForm \cite{Kksal2023LongFormOI} and ELI5 \cite{fan-etal-2019-eli5}. For canonical natural language understanding capabilities, we evaluate all tasks from Open LLM Leaderboard \cite{open-llm-leaderboard} and BBH \cite{suzgun-etal-2023-challenging}. We show performances of models finetuned on full Alpaca 52K and subsets selected various strategies in Table \ref{tab:alpaca_benchmark}. We use METEOR \cite{banerjee-lavie-2005-meteor} score for LongForm and ELI5 as it exhibits higher human correlation following \citet{Kksal2023LongFormOI}. We use exact match for BBH and GSM8K and accuracy for remaining tasks.
While it is not surprising to see the model finetuned with long responses outperforms all other finetuned models on LongForm and ELI5 (\textit{AVG1}), it also achieves better average scores on tasks that do not require long-text generation (\textit{AVG2}). 
Please refer to Appendix Section \ref{sec:benchmark_details} for task details, evaluation details, and results using other SFT datasets.

\section{Conclusion}

Inspired by the superficial nature of SFT, we hypothesize that important SFT data should focus on demonstrations that reflect human-like style, instead of data quality and diversity. We adopt a simple heuristic to select such demonstrations and show its effectiveness compared with various data selection baselines to validate our hypothesis.

\section*{Acknowledgements}

We thank Neeraj Varshney for giving valuable suggestions.

\section*{Limitations}

In this work, to select demonstrations reflecting human-like interactions, we investigate a simple heuristic of selecting instances with long responses, mimicking that detailed responses usually compose detailed contents. However, human-like interactions involve way more features, for example, containing more words showing courtesy and indicating discourse flow, such as "thanks", "sorry", "first of all", and "I mean", etc. We only take a first attempt in this direction by selecting demonstrations with a naive yet effective heuristic. Future directions could focus on designing more effective and accurate strategies to identify human-like demonstrations in various aspects. Also, we only focus on selecting instances from existing SFT datasets; future works could focus on curating SFT datasets from the start or expanding existing SFT datasets to contain more human-like demonstrations, which will likely align LLMs with humans further. Finally, our evaluation heavily relies on GPT-4, which could be biased and costly for extensive usage, a more thorough evaluation would require crowd workers to perform rigorous human evaluation.


\bibliography{custom}

\appendix
\appendix

\section{Related Works}
\noindent \textbf{Instruction Tuning Datasets.} \hspace{4pt}
Instruction tuning has been applied in two main scenarios. One is to improve the general natural language understanding capabilities by multi-task finetuning LLMs on a large collection of NLP tasks, with each task given natural language instructions. Such representative instruction-tuning datasets include Natural Instructions \cite{mishra-etal-2022-cross}, Super-Natural Instructions \cite{wang-etal-2022-super}, OPT-IML \cite{Iyer2022OPTIMLSL}, P3 \cite{sanh2022multitask}, xP3 \cite{muennighoff-etal-2023-crosslingual}, FLAN 2021 \cite{wei2022finetuned}, and FLAN 2022 \cite{Chung2022ScalingIL, 10.5555/3618408.3619349}. Our work focuses on the other scenario, which is to equip LLMs with mainly open-ended instruction-following or conversational capabilities helpfully and harmlessly \cite{Askell2021AGL}. However, collecting open-ended instruction-following data (SFT data) is expensive. One main approach is to distill synthetic data from powerful proprietary LLMs such as ChatGPT and GPT-4; one is to obtain high-quality data completely with human effort; and a final approach is to collect real-world human conversations with SOTA LLMs in a semi-automatic fashion.
Representative open-source distilled datasets include Self-Instruct \cite{wang-etal-2023-self-instruct}, Alpaca 52K \cite{alpaca}, Baize \cite{xu-etal-2023-baize}, UltraChat \cite{ding-etal-2023-enhancing}, Unnatural Instructions \cite{honovich-etal-2023-unnatural}, GPT-4-LLM \cite{Peng2023InstructionTW}, and WizardLM \cite{Xu2023WizardLMEL}.
Representative human-authored open-source datasets include Open Assistant \cite{kopf2023openassistant}, Dolly 15K \cite{DatabricksBlog2023DollyV2}, and LIMA \cite{zhou2023lima}. 
Representative semi-automatic efforts to collect SFT datasets include ShareGPT \footnote{\url{https://sharegpt.com/}}, WildChat \cite{anonymous2024inthewildchat}, and LMSYS-Chat-1M \cite{Zheng2023LMSYSChat1MAL}.

\noindent \textbf{SFT Data Selection.} \hspace{4pt}
As per two important data-centric AI concepts \cite{hajij2021datacentric,10.1145/3580305.3599553}, data quality and diversity contribute significantly to the success of AI systems. High-quality data should be accurate, relevant, and complete to produce reliable results, and models should be trained on diverse datasets to reduce bias \cite{10.1145/2882903.2912574,Motamedi2021ADA}. Recent works have applied these two concepts to the regime of open-ended instruction-tuning. \citet{zhou2023lima} manually curated 1,000 high-quality SFT instances from Stack Exchange and WikiHow, and surprisingly, leading to strong instruction-following models. Following \citet{zhou2023lima}, works have emerged to select subsets of useful instances from a large pool of SFT data to finetune better models compared with utilizing the whole SFT data pool. \citet{Cao2023InstructionMH} propose an instruction mining approach to mine high-quality instances with natural language indicators. 
\cite{Lu2023InsTagIT} utilize ChatGPT to tag SFT instances based on semantics and intentions and select instances with high diversity and high complexity based on those tags. \cite{alpagasus} propose to rate the quality of each SFT instance using ChatGPT and \citet{Du2023MoDSMD, Bukharin2023DataDM} use reward models to assess the quality instead. 
Works \cite{Chen2023MaybeO0, Wu2023SelfEvolvedDD} have been done to select high-diversity instances in instruction embedding space with algorithms like K-Means-Clustering and K-Centering-Sampling \cite{sener2018active}, 
or select based on diversity aspect combined with quality aspect \cite{Bukharin2023DataDM, Du2023MoDSMD}. \citet{Xu2023RethinkingTI} first expand existing SFT dataset based on the diversity aspect with GPT-4, then select high-quality instances from expanded pool.
\citet{Liu2023WhatMG} performs a systematic data selection analysis from diversity, quality, and complexity aspects and combines them all to perform data selection; however, their method also requires expanding existing SFT data with ChatGPT.

Deviating from the above selection methods that directly depend on some quality and diversity metrics, \citet{Li2023FromQT, Li2023OneSL} select useful instances for instruction tuning. \citet{Li2023FromQT} introduce the Instruction-Following Difficulty score to identify the discrepancy between LLM's expected instruction-following capability and its base response capability. \citet{Li2023OneSL} utilizes a scoring system based on the impact of candidate examples on the perplexity of a diverse anchor set, facilitating the selection of the most beneficial data for instruction tuning.

\noindent \textbf{Instruction-Following Evaluation.} \hspace{4pt}
Assessing the instruction-following capabilities of LLMs is challenging due to the open-ended nature of instructions and the expensive cost of performing human evaluations. Researchers have recently started using powerful proprietary LLMs such as GPT-4 as judges \cite{dubois2023alpacafarm, zheng2023judging, vicuna2023} to evaluate a cost-effective alternative. \citet{zheng2023judging, dubois2023alpacafarm} demonstrate that GPT-4 as an evaluator strongly agrees with human evaluations. However, it has also been reported that LLMs as evaluators are sensitive to certain biases such as position bias \cite{Wang2023LargeLM, zheng2023judging}, meaning LLMs might prefer the response that is placed in a specific position (firstly or secondly) during pairwise evaluation, and output length bias, meaning LLMs prefers responses with longer token length. In our work, we mitigate position bias by swapping two responses and let LLM evaluate twice during pairwise comparison. For output length bias, we utilize GPT-4 as the evaluator given that GPT-4 is robust enough to length bias: it only failed 8\% of the time under a repetitive attack, whereas other LLMs judges such as ChatGPT and Claude \cite{claude} fail more than 90\% of the time \cite{zheng2023judging}.

\section{SFT Datasets Details}
\label{sec:datasets_details}
\noindent \textbf{Alpaca 52K.} \hspace{4pt}
A  dataset synthesized by \texttt{text-davinci-003} with a self-instruct-style as done in \citet{wang-etal-2023-self-instruct}; however, with some improvements over the original process. 

\noindent \textbf{WizardLM 70K.} \hspace{4pt} \cite{Xu2023WizardLMEL} construct this dataset leveraging an Evol-Instruct algorithm to rewrite Alpaca 52K step by step by \texttt{gpt-3.5-turbo} into more complex instructions. We use the original sampled 70K subset.

\noindent \textbf{Dolly 15K.} \hspace{4pt} This the first open-source, human-annotated instruction-following dataset \cite{DatabricksBlog2023DollyV2} collected by Databricks employees following the data collection methodology of \citet{NEURIPS2022_b1efde53}.

Besides instruction and response, instances in Alpaca 52K and Dolly 15K also have an input that supplements instruction, if there is any.

\section{Training Details}
\label{sec:training_details}
We use AdamW optimizer \cite{loshchilov2018decoupled} with a learning rate of 2$e-$5 and no weight decay, and a scheduler with linear decay and a warmup ratio of 3\% for the total training steps. We use a batch size of 128 to finetune 3 epochs with a maximum sequence length of 512 for Alpaca, 1,024 for Dolly, and 2,048 for WizardLM. To enable efficient finetuning of LLMs, we use the DeepSpeed library \cite{10.1145/3394486.3406703} and ZeRO stage 3 optimizer \cite{10.5555/3433701.3433727}. All models are trained with BFloat16 \cite{Kalamkar2019ASO} mixed precision for stability. All experiments are conducted on 4 RTX A6000 GPUs.

We use training prompts from FastChat \footnote{\url{https://github.com/lm-sys/FastChat/blob/main/fastchat/conversation.py}} for Alpaca 52K and Dolly 15K. The upper prompt is for instances with input fields, and the lower prompt is for those without input fields:
\begin{tcolorbox}[left=2pt,right=2pt,fontupper=\ttfamily,fontlower=\ttfamily,colback=white,colframe=black,boxrule=1pt]
Below is an instruction that describes a task, paired with an input that provides further context. Write a response that appropriately completes the request.\\ \\
\#\#\# Instruction: \\
$\{$instruction$\}$ \\ \\
\#\#\# Input: \\
$\{$input$\}$ \\ \\
\#\#\# Response:
\tcblower
Below is an instruction that describes a task. Write a response that appropriately completes the request.
\\ \\
\#\#\# Instruction: \\
$\{$instruction$\}$ \\ \\
\#\#\# Response:
\end{tcolorbox}
\noindent We use the original training prompt from \citet{Xu2023WizardLMEL} for WizardLM:
\begin{tcolorbox}[left=2pt,right=2pt,fontupper=\ttfamily,colback=white,colframe=black,boxrule=1pt]
$\{$instruction$\}$ \\ \\ 
\#\#\# Response:
\end{tcolorbox}

\begin{figure*}[!t]
    \centering
    \begin{subfigure}[b]{0.32\textwidth}
     \centering
     \includegraphics[width=1\textwidth]{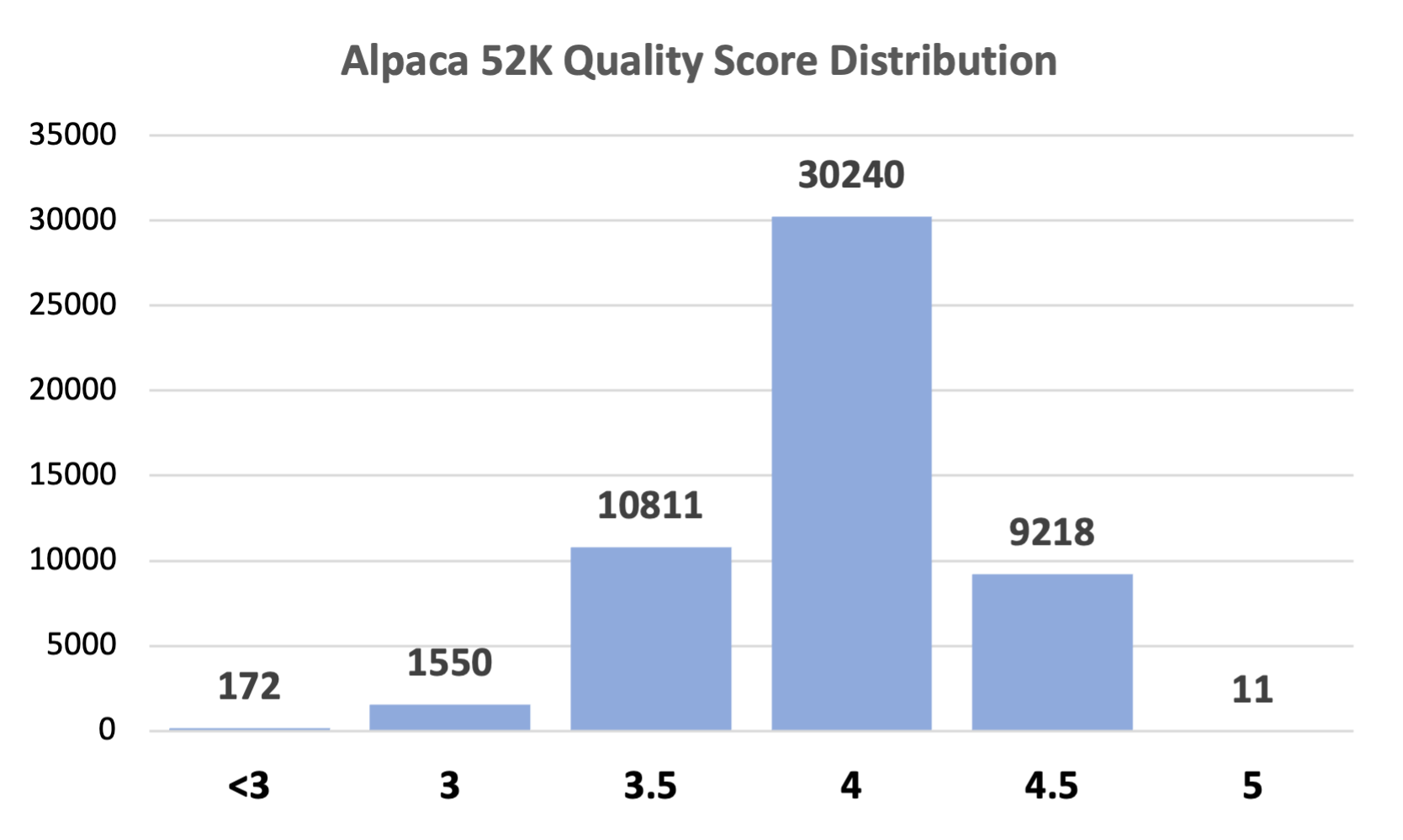}
    \end{subfigure}
    \begin{subfigure}[b]{0.32\textwidth}
     \centering
     \includegraphics[width=1\textwidth]{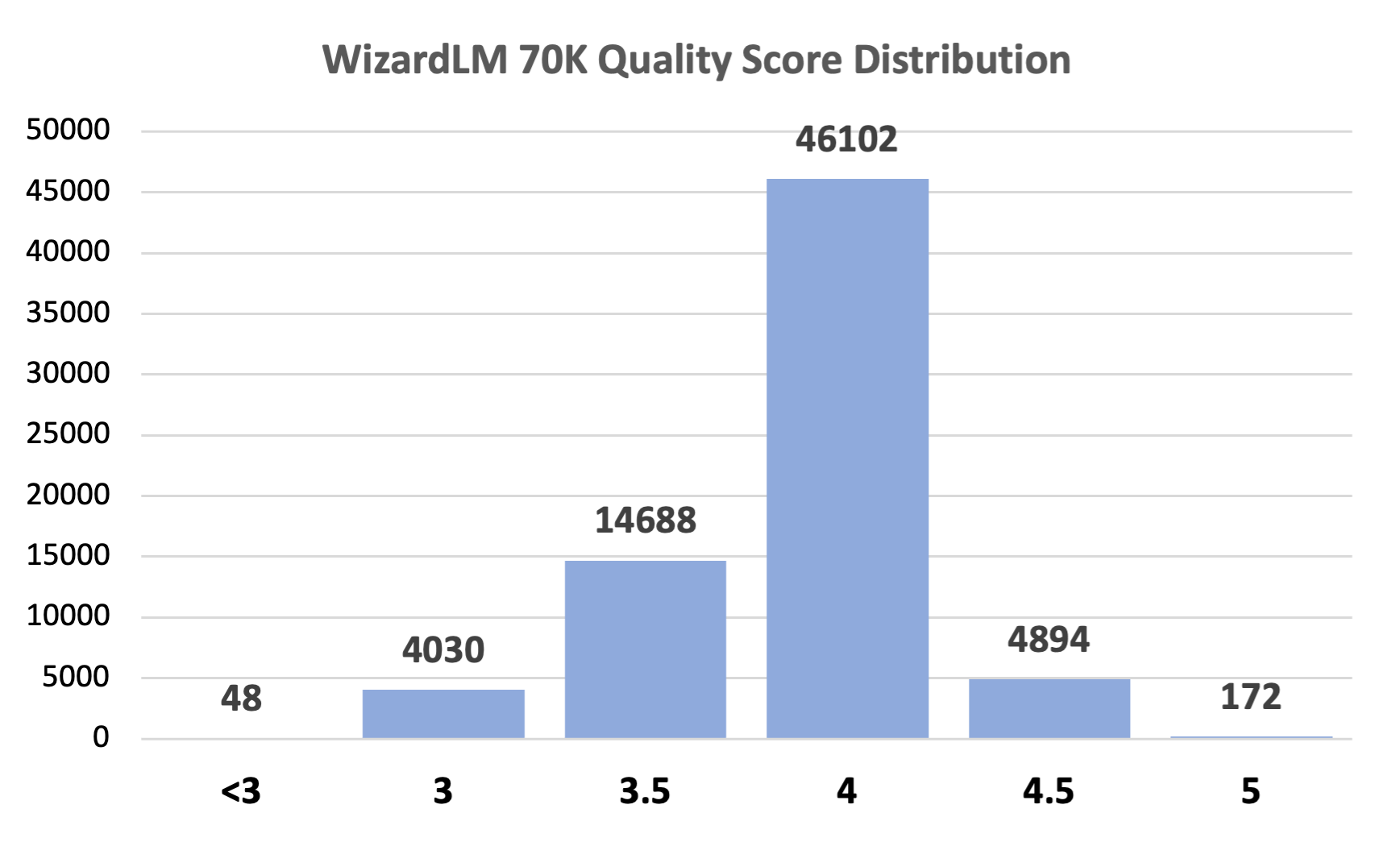}
    \end{subfigure}
    \begin{subfigure}[b]{0.32\textwidth}
     \centering
     \includegraphics[width=1\textwidth]{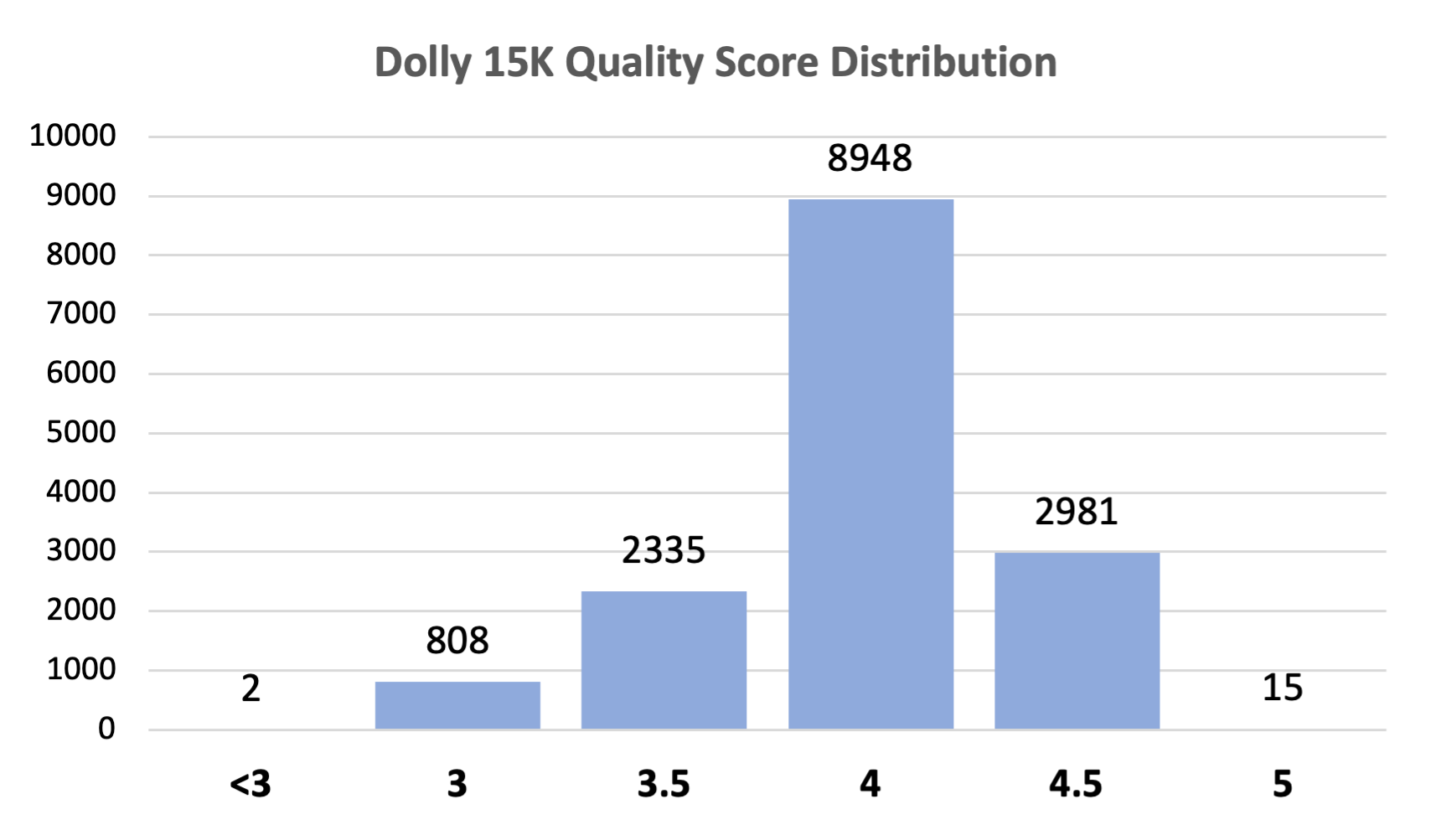}
    \end{subfigure}
     
    \caption{Quality score distributions of the three SFT datasets, rated by ChatGPT.}
    \label{fig:quality_score}
\end{figure*}

\section{High-Quality Selection Details}
\label{sec:quality_selection}
Same as \citet{alpagasus}, we use \texttt{gpt-3.5-turbo-0301} with prompt shown in Figure \ref{fig:quality_prompt} to rate each instance with a quality score from 0 to 5. We show the distributions of scores rated by ChatGPT in Figure \ref{fig:quality_score}. 
We reuse the rating from \citet{alpagasus} for Alpaca 52K and Dolly 15K datasets. Suppose we want to select the top 4,200 high-quality instances from WizardLM dataset with the score distribution shown in the middle subfigure of Figure \ref{fig:quality_score}, we first select all instances with scores of 5, which counts 172 instances, then we randomly sample the remaining 4,028 from instances with scores of 4.5.

\section{Instruction Distribution}
\label{sec:distribution}
\begin{figure}[!t]
    \centering
    \begin{subfigure}[b]{1\linewidth}
        \centering
        \includegraphics[width=0.8\linewidth]{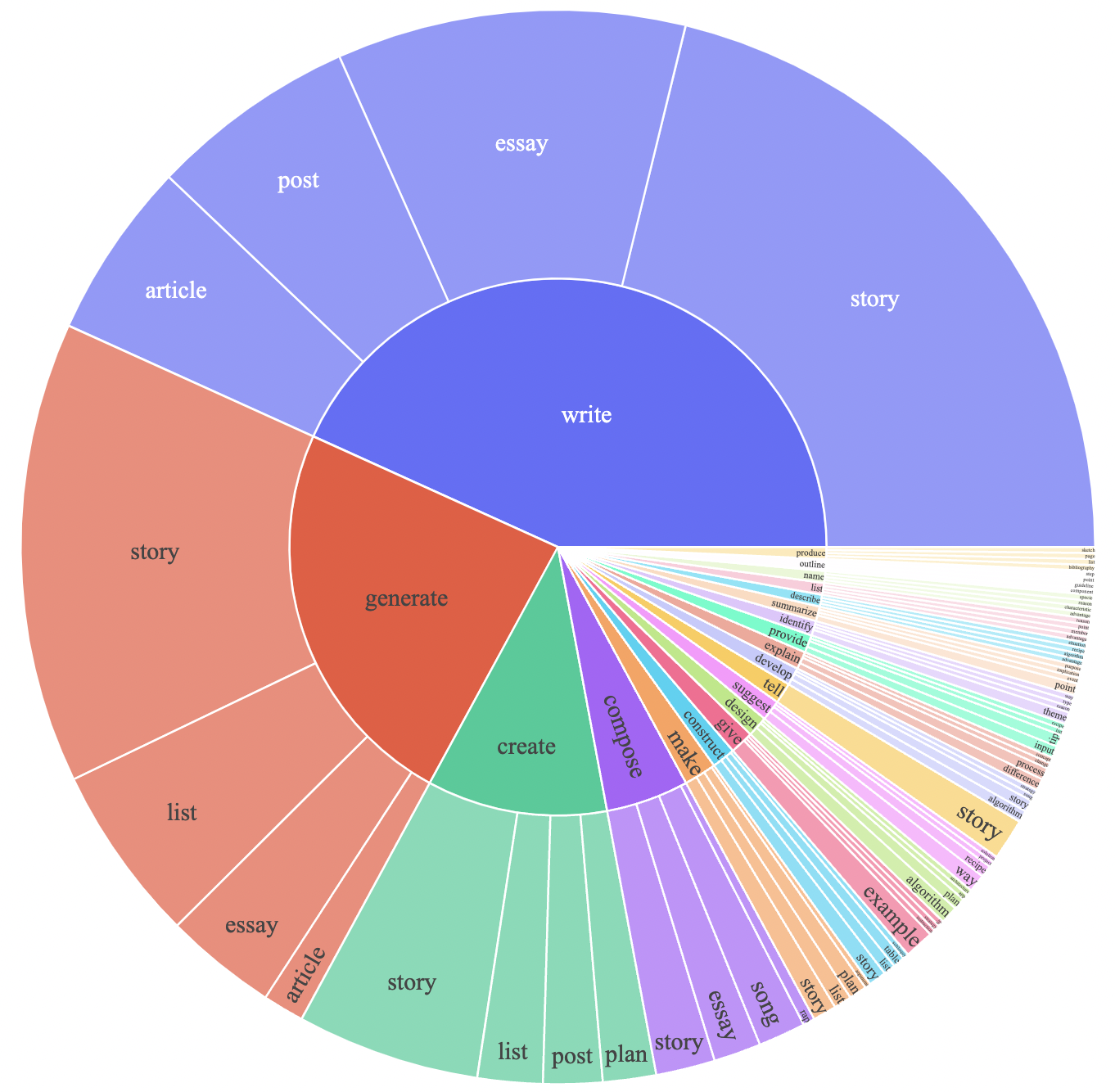}
        \caption{Alpaca-1K}
        \vspace{0.2cm}
    \end{subfigure}
    \begin{subfigure}[b]{1\linewidth}
        \centering
        \includegraphics[width=0.8\linewidth]{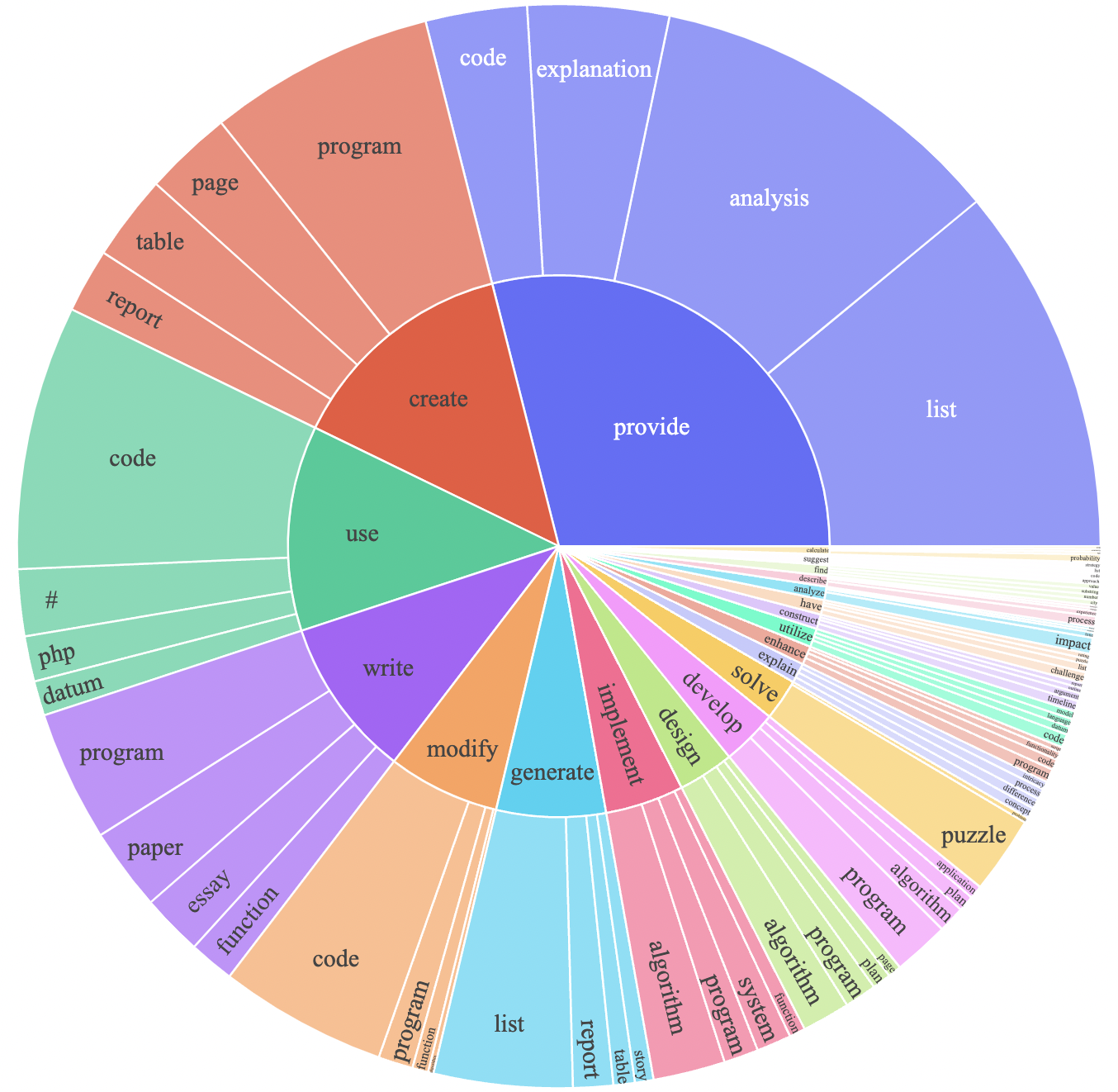}
        \caption{WizardLM-6\%}
    \end{subfigure}
\caption{Verb-noun structures for top instructions with long responses for Alpaca and WizardLM datasets.} 
\label{fig:sunplot}
\end{figure}

We plot the verb-noun structures for instances with long responses from Alpaca and WizardLM in Figure \ref{fig:sunplot}. 
For top 6\% instances with long response in WizardLM 70K, root verbs "\textit{provide}", "\textit{create}", "\textit{use}", "\textit{write}", "\textit{modify}", "\textit{generate}", "\textit{implement}", "\textit{design}", "\textit{develop}", and "\textit{solve}" compose more than 50\% of top instructions, and "\textit{write}", "\textit{generate}", "\textit{create}", and "\textit{compose}" compose 70\% of top 1K instructions from Alpaca. 
We don't plot the distribution of top instances for Dolly since only around one-third of the instructions can be parsed into verb-noun structures by the neural parser. This is possible because Dolly 15K is a human-authored SFT dataset with more complex structures.

\begin{table}[!t]
    \centering
    \scalebox{0.8}{
        \begin{tabular}{@{} cc @{}}
            \toprule
            Test Set & Size \\
            \midrule
            Vicuna & 80 \\
            Koala & 180 \\
            WizardLM & 218 \\
            Self-Instruct & 252 \\
            Lima & 300 \\
            \bottomrule
        \end{tabular}
    }
    
    \caption{Number of instructions in five test sets.}
    \label{tab:testsets_stats}
\end{table}

\begin{table}[!t]
    \centering
    \scalebox{0.72}{
        \begin{tabular}{@{}ccccc@{}}
        \toprule
        \textbf{Dataset} & \textbf{Subset-Size} & \textbf{vs. Full} & \textbf{vs. Random} & \textbf{vs. Longest} \\ \midrule
        \multirow{2}{*}{Alpaca} & Diversity-1K & 0.9650 & 0.8631 & 0.4612 \\
         & Diversity-9K & 1.2029 & 1.0641 & 0.6495 \\ \midrule
        WizardLM & Diversity-6\% & 0.9680 & 0.9379 & 0.8049 \\ \midrule
        \multirow{2}{*}{Dolly} & Diversity-1K & 0.8602 & 1.0583 & 0.7544 \\
         & Diversity-3K & 0.9456 & 1.0621 & 0.8029 \\ \bottomrule
        \end{tabular}
    }
\caption{Winning scores of utilizing high-diversity subset selected by K-Centering-Greedy.} 
\label{tab:coreser}
\end{table}

\begin{table*}[!t]
    \centering
    \scalebox{0.78}{
        \begin{tabular}{@{}cccccccccccc@{}}
        \toprule
        & \textbf{LongForm} & \textbf{ELI5} & \textbf{ARC} & \textbf{HellaSwag} & \textbf{MMLU} & \textbf{TruthfulQA} & \textbf{Winogrande} & \textbf{GSM8K} & \textbf{BBH} & \textbf{\textit{AVG1}} & \textbf{\textit{AVG2}} \\ 
        \midrule
        Base & 14.62 & 11.51 & 50.17 & 72.02 & 45.65 & 38.80 & \textbf{73.80} & 11.50 & 39.63 & 13.07 & 47.37 \\
        \cdashlinelr{1-12}
        Full & 23.38 & 19.42 & 48.38 & 70.68 & 41.40 & 46.15 & 70.48 & 11.50 & 38.98 & 21.40 & 46.80 \\
        Random & 23.57 & 19.74 & \textbf{53.07} & 73.81 & 44.70 & 48.69 & 70.80 & 13.00 & 41.30 & 21.65 & 49.34 \\
        Diversity & 22.93 & 19.67 & 52.22 & 74.15 & 45.90 & \textbf{48.92} & 71.90 & 13.50 & 41.30 & 21.30 & 49.70 \\
        Quality & 24.41 & 20.26 & 52.13 & 73.65 & \textbf{46.60} & 47.67 & 71.51 & \textbf{18.50} & 41.48 & 22.34 & \textbf{50.22} \\
        Longest & \textbf{24.76} & \textbf{20.48} & 51.54 & \textbf{75.40} & 44.20 & 44.84 & 70.56 & 16.00 & \textbf{42.04} & \textbf{22.62} & 49.23 \\
        \bottomrule
        \end{tabular}
    }
    \caption{Performances on all benchmark datasets using base model, models finetuned with full WizardLM dataset, and subsets of it selected with various strategies.}
    \label{tab:wizardlm_benchmark}
\end{table*}

\begin{table*}[!t]
    \centering
    \scalebox{0.78}{
        \begin{tabular}{@{}cccccccccccc@{}}
        \toprule
        & \textbf{LongForm} & \textbf{ELI5} & \textbf{ARC} & \textbf{HellaSwag} & \textbf{MMLU} & \textbf{TruthfulQA} & \textbf{Winogrande} & \textbf{GSM8K} & \textbf{BBH} & \textbf{\textit{AVG1}} & \textbf{\textit{AVG2}} \\ 
        \midrule
        Base & 14.62 & 11.51 & 50.17 & 72.02 & 45.65 & 38.80 & 73.80 & 11.50 & 39.63 & 13.07 & 47.37 \\
        \cdashlinelr{1-12}
        Full & 22.42 & 15.31 & 52.56 & \textbf{76.79} & 45.10 & 41.62 & 71.67 & 10.50 & 39.07 & 18.86 & 48.19 \\
        Random & 20.06 & 14.41 & 54.86 & 75.87 & 45.90 & 39.62 & 74.35 & 12.50 & 32.13 & 17.23 & 47.89 \\
        Diversity & 21.41 & 14.78 & \textbf{55.97} & 75.58 & 46.50 & 43.65 & \textbf{75.30} & 12.50 & 35.37 & 18.10 & 49.27 \\
        Quality & 21.24 & 15.01 & 54.95 & 76.28 & \textbf{47.10} & 42.26 & 74.35 & 13.50 & 35.56 & 18.12 & 49.14 \\
        Longest & \textbf{26.37} & \textbf{18.24} & 54.35 & \textbf{76.79} & 45.80 & \textbf{43.88} & \textbf{75.30} & \textbf{17.00} & \textbf{40.28} & \textbf{22.31} & \textbf{50.49} \\
        \bottomrule
        \end{tabular}
    }
    \caption{Benchmark performances using Dolly 15K as SFT dataset.}
    \label{tab:dolly_benchmark}
\end{table*}

\section{Benchmark Evaluation Details}
\label{sec:benchmark_details}
Six tasks in Open LLM Leaderboard \cite{open-llm-leaderboard} include: ARC-Challenge \cite{Clark2018ThinkYH}, HellaSwag \cite{zellers-etal-2019-hellaswag}, MMLU \cite{hendrycks2021measuring}, TruthfulQA \cite{lin-etal-2022-truthfulqa}, Winogrande \cite{10.1145/3474381}, and GSM8K \cite{Cobbe2021TrainingVT}.

\noindent \textbf{LongForm.} \hspace{4pt} 
The LongForm dataset is created by first leveraging existing corpora such as C4 and Wikipedia with augmented instructions generated via LLMs, then extended with structured corpora examples such as Stack Exchange and WikiHow and task examples such as question answering, email writing, grammar error correction, story/poem generation, and text summarization. We use its test set for evaluation.

\noindent \textbf{ELI5.} \hspace{4pt}
ELI5 is a dataset for long-form question answering gathered from the Reddit forum “Explain Like I’m Five” (ELI5), where users ask factual questions requiring paragraph-length or longer answers. For each question, multiple reference answers are given, and we use the reference with the most votes as the ground truth reference. We use its validation set for evaluation.

\noindent \textbf{BBH.} \hspace{4pt}
BIG-Bench Hard is a suite of 23 challenging BIG-Bench tasks \cite{srivastava2023beyond} for which prior language model evaluations did not outperform the average human-rater. Following the original setup of \citet{suzgun-etal-2023-challenging}, we use 3 few-shot in-context examples with chain-of-thought prompting \cite{wei2022chain} during inference. We randomly sample 50 examples per task for faster evaluation following \citet{wang2023how}.

\noindent \textbf{ARC.} \hspace{4pt}
We use the Challenge subset from AI2 Reasoning Challenge, a set of grade-school level, multiple-choice science questions. We use its test set with 25-shot examples during inference following \citet{open-llm-leaderboard}. 

\noindent \textbf{HellaSwag.} \hspace{4pt}
HellaSwag is a challenging dataset for studying grounded commonsense inference. It was hard for state-of-the-art models at the time of release, though trivial for humans. We use its validation set with 10-shot examples during inference following \citet{open-llm-leaderboard}. 

\noindent \textbf{MMLU.} \hspace{4pt}
Massive Multitask Language Understanding is a benchmark designed to measure knowledge acquired during pretraining by evaluating models exclusively in zero-shot and few-shot settings. We use the official MMLU evaluation script and prompts available with 5 few-shot examples\footnote{\url{https://github.com/hendrycks/test}} on its test set for evaluation.  

\noindent \textbf{TruthfulQA.} \hspace{4pt}
TruthfulQA is a test to measure a model's propensity to reproduce falsehoods commonly found online. We use the multiple-choice setting of TruthfulQA. We use the default QA prompt format with 6 in-context QA examples.

\noindent \textbf{Winogrande.} \hspace{4pt}
Winograndeis an adversarial and difficult Winograd benchmark at scale for commonsense reasoning. We use its validation set with 5-shot examples during inference following \citet{open-llm-leaderboard} for evaluation.

\noindent \textbf{GSM8K.} \hspace{4pt}
GSM8K is a diverse grade school math word problems dataset to measure a model's ability to solve multi-step mathematical reasoning problems. We evaluate 200 examples randomly sampled from its test set with 8 few-shot in-context examples with chain-of-thought prompting following \citet{wei2022chain} for faster evaluation \cite{wang2023how}.

Here, we present results for data selecting from WizardLM 70K and Dolly 15K in Tables \ref{tab:wizardlm_benchmark} and \ref{tab:dolly_benchmark} (See Table \ref{tab:alpaca_benchmark} for Alpaca 52K). For long-form generation tasks, it is not surprising to see finetuning with long responses leads to the best-performing model. For canonical language understanding tasks, we observe that model finetuned with long responses outperforms all other models on Dolly 15K, same as the trend we observe for Alpaca 52K. However, for WizardLM 70K, although model finetuned with long responses is not the best performing one, it only lags behind the best model by 1\%.

\begin{figure*}
    \centering

    \begin{subfigure}[b]{1\textwidth}
        \begin{tcolorbox}[left=7pt,right=7pt,colback=white,colframe=black,boxrule=1pt,fontupper=\ttfamily]
            \textbf{System Prompt}: \\
            We would like to request your feedback on the performance of AI assistant in response to the instruction displayed following. \\ \\
            Instruction: \{Instruction\} \\
            Response: \{Response\} \\ \\
            \textbf{User Prompt}: \\
            Please rate according to the accuracy of the response to the instruction. Each assistant receives a score on a scale of 0 to 5, where a higher score indicates higher level of the accuracy. Please first output a single line containing value indicating the scores. In the subsequent line, please provide a comprehensive explanation of your evaluation, avoiding any potential bias.
        \end{tcolorbox}
    \caption{\small Prompt for instances with no input.}
    \end{subfigure}

    \vspace{1em}
    
    \begin{subfigure}[b]{1\textwidth}
       \begin{tcolorbox}[left=7pt,right=7pt,colback=white,colframe=black,boxrule=1pt,fontupper=\ttfamily]
            \textbf{System Prompt}: \\
            We would like to request your feedback on the performance of AI assistant in response to the instruction and the given input displayed following. \\ \\
            Instruction: \{Instruction\} \\
            Input: \{Input\} \\
            Response: \{Response\} \\ \\
            \textbf{User Prompt}: \\
            Please rate according to the accuracy of the response to the instruction and the input. Each assistant receives a score on a scale of 0 to 5, where a higher score indicates higher level of the accuracy. Please first output a single line containing value indicating the scores. In the subsequent line, please provide a comprehensive explanation of your evaluation, avoiding any potential bias.
        \end{tcolorbox}
    \caption{\small Prompt for instances with input.}
    \end{subfigure}
    
    \caption{\small The prompt we use to prompt ChatGPT for quality score.} 
    \label{fig:quality_prompt}
\end{figure*}

\begin{figure*}
    \centering
    \begin{subfigure}[b]{1\textwidth}
        \begin{tcolorbox}[left=7pt,right=7pt,colback=white,colframe=black,boxrule=1pt,fontupper=\ttfamily]
            \textbf{System Prompt}: \\
            You are a helpful and precise assistant for checking the quality of the answer. \\ \\
            \textbf{User Prompt}: \\
            \begin{bracket}[Question]\end{bracket}  \\ 
            $\{$question$\}$ \\ \\
            \begin{bracket}[The Start of Assistant 1's Answer]\end{bracket} \\
            $\{$response\_1$\}$ \\
            \begin{bracket}[The End of Assistant 1's Answer]\end{bracket} \\ \\
            \begin{bracket}[The Start of Assistant 2's Answer]\end{bracket} \\
            $\{$response\_2$\}$ \\
            \begin{bracket}[The End of Assistant 2's Answer]\end{bracket} \\ \\
            \begin{bracket}[System]\end{bracket} \\
            We would like to request your feedback on the performance of two AI assistants in response to the user question displayed above. \\
            Please rate the helpfulness, relevance, accuracy, level of details of their responses. Each assistant receives an overall score on a scale of 1 to 10, where a higher score indicates better overall performance. \\
            Please first output a single line containing only two values indicating the scores for Assistant 1 and 2, respectively. The two scores are separated by a space. In the subsequent line, please provide a comprehensive explanation of your evaluation, avoiding any potential bias and ensuring that the order in which the responses were presented does not affect your judgment.
        \end{tcolorbox}
    \end{subfigure}
    \caption{\small The prompt we use for GPT-4 evaluation.} 
    \label{fig:gpt-prompt}
\end{figure*}

\begin{table*}[!h]
    \centering
    \scalebox{0.9}{
        \begin{tabular}{p{1\linewidth}}
             \toprule
             \textbf{Example 1} \\
             \toprule
             \textbf{Instruction}: \newline
              \texttt{Categorize the given magazine based on its content into one of the categories: Politics, Entertainment News, Tech News, Business News, Parenting \& Kids, Health \& Healing, Cooking, Food \& Wine, and Arts \& Literature.} \newline
             \textbf{Input}: \newline
             \texttt{Inc.} \newline \newline
             \textbf{Assistant 1 Response}: \newline
             \texttt{Business News} \newline 
             \textbf{Assistant 2 Response}: \newline
             \texttt{Inc. is a business magazine that focuses on small businesses and entrepreneurship. It provides information on topics such as financing, marketing, management, and legal issues. It also features stories about successful entrepreneurs and their businesses. Inc. also covers topics such as technology, politics, and culture. It is published monthly and has a circulation of over 3 million.} \newline \newline
             \textbf{GPT-4 Evaluation}: \newline
             \texttt{
                7 9 \newline
                Assistant 1 provided a detailed and accurate description of the magazine, Inc., correctly identifying it as a business magazine. The assistant also provided additional information about the magazine's content, publication frequency, and circulation, which was not asked for but adds to the richness of the response. Therefore, Assistant 1 receives a score of 9. \newline
                Assistant 2 correctly categorized the magazine as Business News, which is accurate and directly answers the question. However, the response lacks any additional detail or context about the magazine, which could have made the response more informative and helpful. Therefore, Assistant 2 receives a score of 7.
             } \\
             \bottomrule
        \end{tabular}
    }
    \caption{An example of two responses both meet the instruction requirement; however, GPT-4 prefers the longer one (Assistant 2) given its detailedness. Assistant 1 is LLaMA-2-7B finetuned with entire Alpaca dataset, and Assistant 2 is LLaMA-2-7B finetuned with top 1K instances from Alpaca with long responses.}
    \label{tab:gpt4_explaination_1}
\end{table*}

\begin{table*}[!h]
    \centering
    \scalebox{0.9}{
        \begin{tabular}{p{1\linewidth}}
             \toprule
             \textbf{Example 2} \\
             \toprule
             \textbf{Instruction}: \newline
              \texttt{Think of topics that are most common in classic interview questions for a job in computer science.} \newline \newline 
             \textbf{Assistant 1 Response}: \newline
             \texttt{Common topics in classic interview questions for a job in computer science include algorithms, data structures, programming languages, computer architecture, and software engineering.} \newline 
             \textbf{Assistant 2 Response}: \newline
             \texttt{Common topics in classic interview questions for a job in computer science include: \newline
                - Data Structures and Algorithms \newline
                - Programming Languages \newline
                - Software Development Life Cycle \newline
                - Database Management Systems \newline
                - Computer Networks \newline
                - Operating Systems \newline
                - Cryptography \newline
                - Artificial Intelligence \newline
                - Machine Learning \newline
                - Theoretical Computer Science \newline
                - Computer Architecture \newline
                - Ethical Hacking \newline
                - Cybersecurity \newline
                - Big Data \newline
                - Cloud Computing \newline
                - Internet of Things \newline
                - Blockchain \newline
                - Game Development \newline
                - Web Development \newline
                - Mobile App Development \newline
                - Virtual Reality \newline
                - Augmented Reality \newline
                - Robotics \newline
                - Embedded Systems \newline
                - Networking \newline
                - Hardware Design \newline
                - Troubleshooting \newline
                - Project Management \newline
                - Professional Ethics} \newline \newline
             \textbf{GPT-4 Evaluation}: \newline
             \texttt{
                7 10 \newline
                Assistant 1 provided a correct and relevant answer, but it was quite brief and lacked detail. The assistant mentioned only five topics, which are indeed common in computer science interviews, but there are many more topics that could be included. Therefore, the assistant gets a score of 7. \newline
                Assistant 2, on the other hand, provided a very comprehensive list of topics that are common in computer science interviews. The assistant included a wide range of topics, from basic ones like data structures and algorithms, to more specialized ones like artificial intelligence, machine learning, and cybersecurity. The assistant also mentioned topics related to the practical aspects of the job, like project management and professional ethics. This answer is not only accurate and relevant, but also very detailed, which makes it very helpful for someone preparing for a computer science job interview. Therefore, Assistant 2 gets a perfect score of 10.
             } \\
             \bottomrule
        \end{tabular}
    }
    \caption{Second example, continued from Table \ref{tab:gpt4_explaination_1}.}
    \label{tab:gpt4_explaination_2}
\end{table*}


\end{document}